\documentclass{book} 
\usepackage{amsmath,latexsym,amssymb,kcp,named} 
\usepackage[latin1]{inputenc}
\usepackage{epsfig}
\usepackage{tabularx}
\collection

\newcolumntype{C}{>{$}c<{$}}
\newcolumntype{L}{>{$}l<{$}}
\newcolumntype{R}{>{$}r<{$}}

\begin{document}


\paper[Neural-symbolic Integration] {Dimensions of Neural-symbolic\protect
  \\[.5ex] Integration --- A Structured Survey} { Sebastian
  Bader\thanks{Supported by the German Research Foundation (DFG) under GK334.}
  and Pascal Hitzler\thanks{Supported by the German Federal Ministry of
    Education and Research (BMBF) under the SwartWeb project.}}

\section{Introduction}

Research on integrated neural-symbolic systems has made significant progress in
the recent past. In particular the understanding of ways to deal with symbolic
knowledge within connectionist systems (also called artificial neural networks)
has reached a critical mass which enables the community to strive for applicable
implementations and use cases.  Recent work has covered a great variety of
logics used in artificial intelligence and provides a multitude of techniques
for dealing with them within the context of artificial neural networks.
        
Already in the pioneering days of computational models of neural cognition, the
question was raised how symbolic knowledge can be represented and dealt with
within neural networks. The landmark paper \cite{CP43} provides fundamental
insights how propositional logic can be processed using simple artificial neural
networks. Within the following decades, however, the topic did not receive much
attention as research in artificial intelligence initially focused on purely
symbolic approaches. The power of machine learning using artificial neural
networking was not recognized until the 80s, when in particular the
backpropagation algorithm \cite{RHW86} made connectionist learning feasible and
applicable in practice.
        
These advances indicated a breakthrough in machine learning which quickly led to
industrial-strength applications in areas such as image analysis, speech and
pattern recognition, investment analysis, engine monitoring, fault diagnosis,
etc.  During a training process from raw data, artificial neural networks
acquire expert knowledge about the problem domain, and the ability to generalize
this knowledge to similar but previously unencountered situations in a way which
often surpasses the abilities of human experts.  The knowledge obtained during
the training process, however, is hidden within the acquired network
architecture and connection weights, and not directly accessible for analysis,
reuse, or improvement, thus limiting the range of applicability of the neural
networks technology. For these purposes, the knowledge would be required to be
available in structured symbolic form, most preferably expressed using some
logical framework.
        
Likewise, in situations where partial knowledge about an application domain is
available before the training, it would be desirable to have the means to guide
connectionist learning algorithms using this knowledge. This is the case in
particular for learning tasks which traditionally fall into the realm of
symbolic artificial intelligence, and which are characterized by complex and
often recursive interdependencies between symbolically represented pieces of
knowledge.
        
The arguments just given indicate that an integration of connectionist and
symbolic approaches in artificial intelligence provides the means to address
machine learning bottlenecks encountered when the paradigms are used in
isolation. Research relating the paradigms came into focus when the limitations
of purely connectionist approaches became apparent. The corresponding research
turned out to be very challenging and produced a multitude of very diverse
approaches to the problem. Integrated systems in the sense of this survey are
those where symbolic processing functionalities emerge from neural structures
and processes.
                
\begin{figure}
  \centering
  \begin{minipage}{\linewidth}
    \epsfig{figure=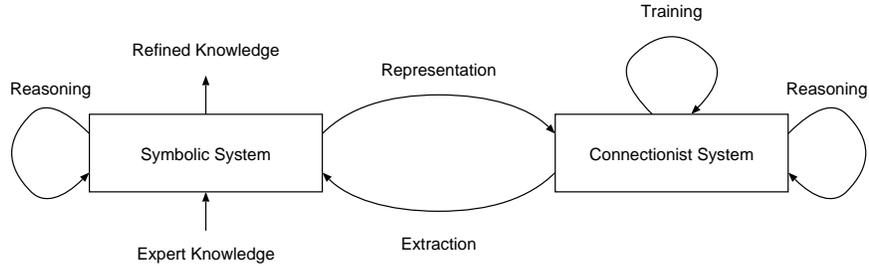, width=\linewidth}
  \end{minipage}
  \caption{Neural-symbolic learning cycle}
  \label{fig:nesycycle}
\end{figure}

Most of the work in integrated neural-symbolic systems addresses the
neural-symbolic learning cycle depicted in Figure \ref{fig:nesycycle}.
A front-end (symbolic system) is used to feed symbolic (partial) expert
knowledge to a neural or connectionist system which can be trained on raw data,
possibly taking the internally represented symbolic knowledge into account.
Knowledge acquired through the learning process can then be extracted back to
the symbolic system (which now also acts as a back-end), and made available for
further processing in symbolic form. Studies often address only parts of the
neural-symbolic learning cycle (like the representation or extraction of
knowledge), but can be considered to be part of the overall investigations
concerning the cycle.

We assume that the reader has a basic familiarity with artificial neural
networks and symbolic artificial intelligence, as conveyed by any introductory
courses or textbooks on the topic, e.g. in \cite{russell:norvig:03}.  However,
we will refrain from going into technical detail at any point, but rather
provide ample references which can be followed up at ease.  The selection of
research results which we will discuss in the process is naturally subjective
and driven by our own specific research interests.  Nevertheless, we hope that
this survey also provides a helpful and comprehensive albeit unusual literature
overview to neural-symbolic integration.

This chapter is structured as follows. In Section \ref{sec:nesysystems}, we
introduce some of those integrated neural-symbolic systems, which we consider to
be foundational for the majority of the work undertaken within the last decade.
In Section \ref{sec:classification}, we will explain our proposal for a
classification scheme. In Section \ref{sec:dimensions}, we will survey recent
literature by means of our classification. Finally, in Section
\ref{sec:conclusions}, we will give an outlook on possible further developments.

\section{Neural-Symbolic Systems}
\label{sec:nesysystems}

As a reference for later sections, we will review some well-known systems here.
We will start with the landmark results by McCulloch and Pitts, which relate
finite automata and neural networks \cite{CP43}. Then we will discuss a method
for representing structured terms in a connectionist systems, namely the
recursive autoassociative memories (RAAM) \cite{pollack:90}.  The SHRUTI System,
proposed in \cite{SA93}, is discussed next.  Finally, \emph{Connectionist Model
  Generation using the Core Method} is introduced as proposed in \cite{HK94}.
These approaches lay the foundations for most of the more recent work on
neural-symbolic integration which we will discuss in this chapter.

\subsection{Neural Networks and Finite Automata}
\label{sec:cp43}

The advent of automata theory and of artificial neural networks, marked also the
advent of neural-symbolic integration. In their seminal paper \cite{CP43} Warren
Sturgis McCulloch and Walter Pitts showed that there is a strong relation
between symbolic systems and artificial neural networks. In particular, they
showed that for each finite state machine there is a network constructed from
binary threshold units -- and vice versa -- such that the input-output behaviour
of both systems coincide. This is due to the fact that simple logical
connectives such as conjunction, disjunction and negation can easily be encoded
using binary threshold units, with weights and thresholds set appropriately.  To
illustrate the ideas, we will discuss a simple example in the sequel.

\begin{example}
  Figure~\ref{fig:CP43} on the left shows a simple Moore-machine, which is a
  finite state machine with outputs attached to the states
  \cite{Hopcroft:Ullman:1979}.  The corresponding network is shown on the right.
  The network consists of four layers. For each output-symbol (${0, 1}$) there
  is a unit in the output-layer, and for each input-symbol (${a, b}$) a unit in
  the right part of the input-layer.  Furthermore, for each state (${q_0, q_1}$)
  of the automaton, there is a unit in the state-layer and in the left part of
  the input layer. In our example, there are two ways to reach the state $q_1$,
  namely by being in state $q_1$ and receiving an '$a$', or by being in state
  $q_0$ and receiving a '$b$'. This is implemented by using a disjunctive neuron
  in the state-layer receiving inputs from two conjunctive units in the gate
  layer, which are connected to the corresponding conditions, as e.g. being in
  state $q_0$ and reading a '$b$'.
  \begin{figure}
    \centering
    \begin{minipage}{.4\linewidth}
      \epsfig{figure=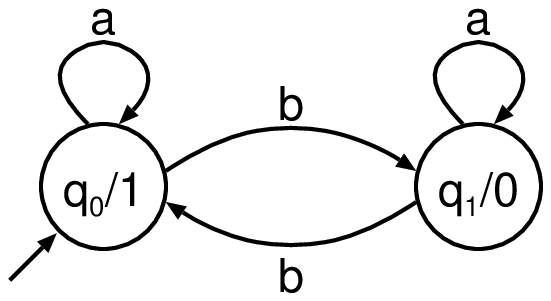, width=\linewidth}
      \footnotesize
      \begin{align*}
        q_0/1&\stackrel{a}{\longrightarrow}q_0/1\stackrel{b}{\longrightarrow}q_1/0\\
        &\stackrel{b}{\longrightarrow} q_0/1\stackrel{a}{\longrightarrow}q_0/1
      \end{align*}
    \end{minipage}
    \hspace{.1\linewidth}
    \begin{minipage}{.4\linewidth}
      \epsfig{figure=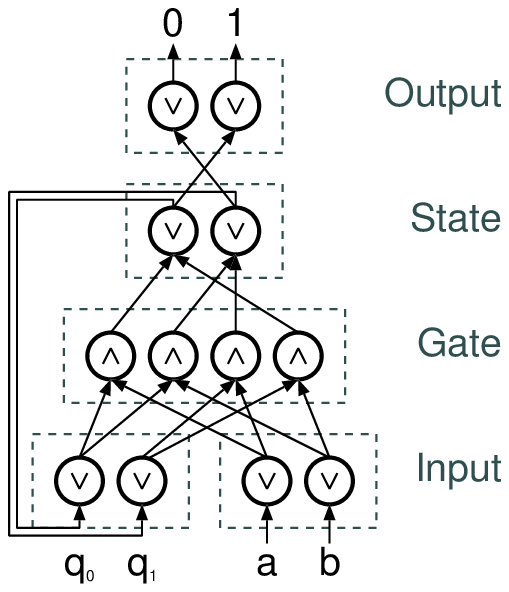, width=\linewidth}
    \end{minipage}
    \caption{A simple Moore-machine, the processing of the input word $abba$ (left)
      and a corresponding McCulloch-Pitts-network (right).}
    \label{fig:CP43}
  \end{figure}
\end{example}

A network of $n$ binary threshold units can be in $2^n$ different states only,
and the change of state depends on the current input to the network only. These
states and transitions can easily be encoded as a finite automaton, using a
straightforward translation \cite{CP43,KleeneAS56}. An extension to the class of
weighted automata is given in \cite{BHS04}.

\subsection{Connectionist Term Representation}
\label{sec:raams}

The representation of possibly infinite structures in a finite network is one of
the major obstacles on the way to neural-symbolic integration \cite{BHH04}. One
attempt to solve this will be discussed in this section, namely the idea of
\emph{recursive autoassociative memories (RAAMs)} as introduced in
\cite{pollack:90}, where a fixed length representation of variable sized data is
obtained by training an artificial neural network using backpropagation. Again,
we will try to illustrate the ideas by discussing a simple example.

\begin{example}
  Figure~\ref{fig:raam} shows a small binary tree which shall be encoded in a
  fixed-length real vector. The resulting RAAM-network is depicted in
  Figure~\ref{fig:raam}, where each box depicts a layer of 4 units. The network
  is trained as an encoder-decoder network, i.e. it reproduces the input
  activations in the output layer \cite{Bis95}.  In order to do this, it must
  create a compressed representation in the hidden layer.
  Table~\ref{tab:raam:samples} shows the activations of the layers during the
  training of the network. As the training converges we shall have $A=A'$,
  $B=B'$, etc.  To encode the terminal symbols $A$, $B$, $C$ and $D$ we use the
  vectors $(1,0,0,0)$, $(0,1,0,0)$, $(0,0,1,0)$ and $(0,0,0,1)$ respectively.
  The representations of $R_1$, $R_2$ and $R_3$ are obtained during training.
  After training the network, it is sufficient to keep the internal
  representation $R_3$, since it contains all necessary information for
  recreating the full tree. This is done by plugging it into the hidden layer
  and recursively using the output activations, until binary vectors, hence
  terminal symbols, are reached.

  \begin{table}
    \centering
    \begin{tabular}{CCCCC}
      \text{\bf Input}  &           &  \text{\bf Hidden}  &&  \text{\bf Output} \\
      \hline
      (A\ B)            &\rightarrow&  R_1(t)  &\rightarrow&  (A'(t)\ B'(t))    \\
      (C\ D)            &\rightarrow&  R_2(t)  &\rightarrow&  (C'(t)\ D'(t))    \\
      (R_1(t)\ R_2(t))  &\rightarrow&  R_3(t)  &\rightarrow&  (R_1'(t)\ R_2'(t))
    \end{tabular}
    \caption{Extracted training samples from the tree shown in Figure~\ref{fig:raam}.}
    \label{tab:raam:samples}
  \end{table}

  \begin{figure}
    \centering
    \begin{minipage}{.45\linewidth}
      \epsfig{figure=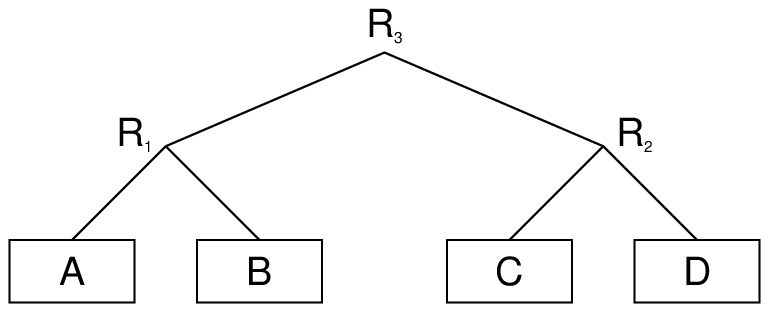, width=\linewidth}
    \end{minipage}
    \hspace{.1\linewidth}
    \begin{minipage}{.35\linewidth}
      \epsfig{figure=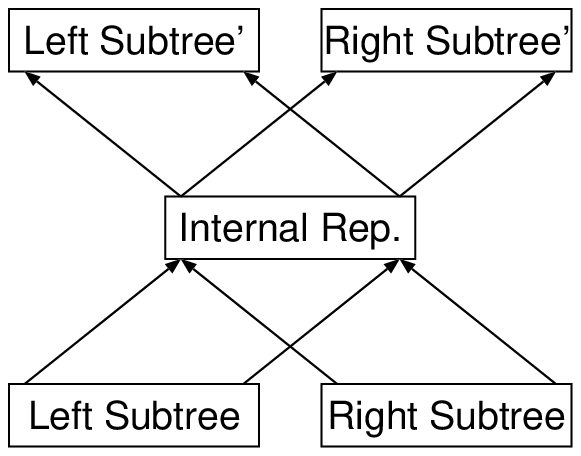, width=\linewidth}
    \end{minipage}
    \caption{Example tree and a RAAM for binary trees.}
    \label{fig:raam}
  \end{figure}
\end{example} 

While recreating the tree from its compressed representation, it is necessary to
distinguish terminal and non-terminal vectors, i.e. those which represent leafs
of the trees from those representing nodes. Due to noise or inaccuracy, it can
be very hard to recognise the ``1-of-n''-vectors representing terminal symbols.
In order to circumvent this problem different solutions were proposed, which can
be found in \cite{Stolcke:Wu:1992,sperduti:93,sperduti:93:a}.  The ideas
described above for binary vectors apply also for trees with larger, but fixed,
branching factors, by simply using bigger input and output layers. In order to
store sequences of data, a version called S-RAAM (for sequential RAAM) can be
used \cite{pollack:90}.  In \cite{Blair:1997} modifications were proposed to
allow the storage of deeper and more complex data structures than before, but
their applicability remains to be shown \cite{kalinke:97}. Other recent
approaches for enhancement have been studied e.g. in
\cite{SSG95,KwasnyKalman95,SSG97,Hammerton98phd,AD99}, which also include some
applications. A recent survey which includes RAAM architectures and addresses
structured processing can be found in \cite{FGKS01sequences}. The related
approach on \emph{Holographic reduced representations (HRRs)}
\cite{Plate:1991,Plate:1995} also uses fixed-length representations of
variable-sized data, but using different methods.

\subsection{Reflexive Connectionist Reasoning}
\label{sec:shruti}

A wide variety of tasks can be solved by humans very fast and efficiently. This
type of reasoning is sometimes referred to as \emph{reflexive reasoning}.  The
SHRUTI system \cite{SA93} provides a connectionist architecture performing this
type of reasoning.  Relational knowledge is encoded by clusters of cells and
inferences by means of rhythmic activity over the cell clusters. It allows to
encode a (function-free) fragment of first-order predicate logic analyzed in
\cite{HKW99}. Binding of variables -- a particularly difficult aspect of
neural-symbolic integration -- is obtained by time-synchronization of activities
of neurons.

\begin{example}
  Table~\ref{table:shruti:kb} shows a knowledge base describing what it means to
  own something and to be able to sell it. Furthermore it states some facts. The
  resulting SHRUTI network is shown in Figure~\ref{fig:shruti}.
 
  \begin{table}
    \centering
    \begin{tabular}{L@{\qquad}L}
      \text{\bf Rules}                      & \text{\bf Facts} \\\hline
      Owns(y,z) \gets Gives(x,y,z)          & Gives(john, josephine, book) \\
      Owns(x,y) \gets Buys(x,y)             & Buys(carl, x) \\
      Can\!\!-\!\!sell(x,y) \gets Owns(x,y) & Owns(josephine, ball)
    \end{tabular}
    \caption{A knowledge base for $Own$ and $Can\!\!-\!\!sell$}
    \label{table:shruti:kb}
  \end{table}

  \begin{figure}
    \centering
    \begin{minipage}{.9\linewidth}
      \epsfig{figure=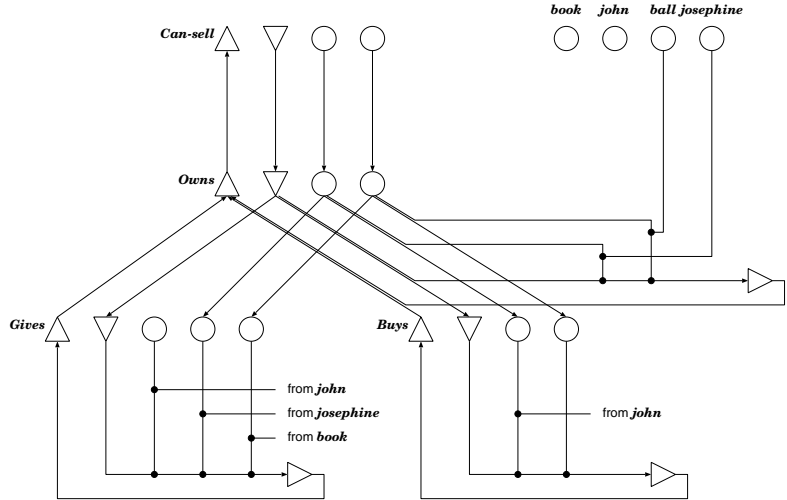, width=\linewidth}
    \end{minipage}
    \caption{A SHRUTI network for the knowledge base from Table~\ref{table:shruti:kb}.
      Each predicate is represented by two \emph{relais units} ($\bigtriangleup,
      \bigtriangledown$) and a set of \emph{argument units} ($\bigcirc$).
      Constants are represented as $\bigcirc$-units in the upper right. Facts
      are implemented using $\triangleright$-units.}
    \label{fig:shruti}
  \end{figure}
\end{example}

Recent enhancements, as reported in \cite{Sha99} and \cite{SW99}, allow e.g. the
support of negation and inconsistency. \cite{WS03pcs} adds very basic learning
capabilities to the system, while \cite{WS04cs} addresses the problem of
multiple reuse of knowledge rules, an aspect which limits the capabilities of
SHRUTI.

\subsection{Connectionist Model Generation using the Core Method}\label{sec:HK94etc}

In 1994, Hölldobler and Kalinke proposed a method to translate a propositional
logic program into a neural network \cite{HK94} (a revised treatment is
contained in \cite{HHS0xjal}), such that the network will settle down in a state
corresponding to a model of the program. To achieve this goal, not the program
itself, but rather the associated consequence operator was implemented using a
connectionist system. The realization is close in spirit to \cite{CP43}, and
Figure~\ref{fig:hk94} shows a propositional logic program and the corresponding
network.

\begin{example}
  The simple logic program in Figure~\ref{fig:hk94} states that $a$ is a fact,
  $b$ follows from $a$, etc. This ``follows-from'' is usually captured by the
  associated consequence operator $T_P$ \cite{Llo88}. The figure shows also the
  corresponding network, obtained by the algorithm given in \cite{HK94}. For
  each atom ($a,b,c,d,e$) there is a unit in the input- and output layer, whose
  activation represents the truth value of the corresponding atom. Furthermore,
  for each rule in the program there is a unit in the hidden layer, acting as a
  conjunction. If all requirements are met, this unit becomes active and
  propagates its activation to the consequence-unit in the output layer.

  \begin{figure}
    \centering
    \begin{minipage}{.3\linewidth}
      $C_1: a.$ \\
      $C_2: b \gets a.$ \\
      $C_3: c \gets a, b$\\
      $C_4: d \gets e$\\
      $C_5: e \gets d$
    \end{minipage}
    \begin{minipage}{.5\linewidth}
      \epsfig{figure=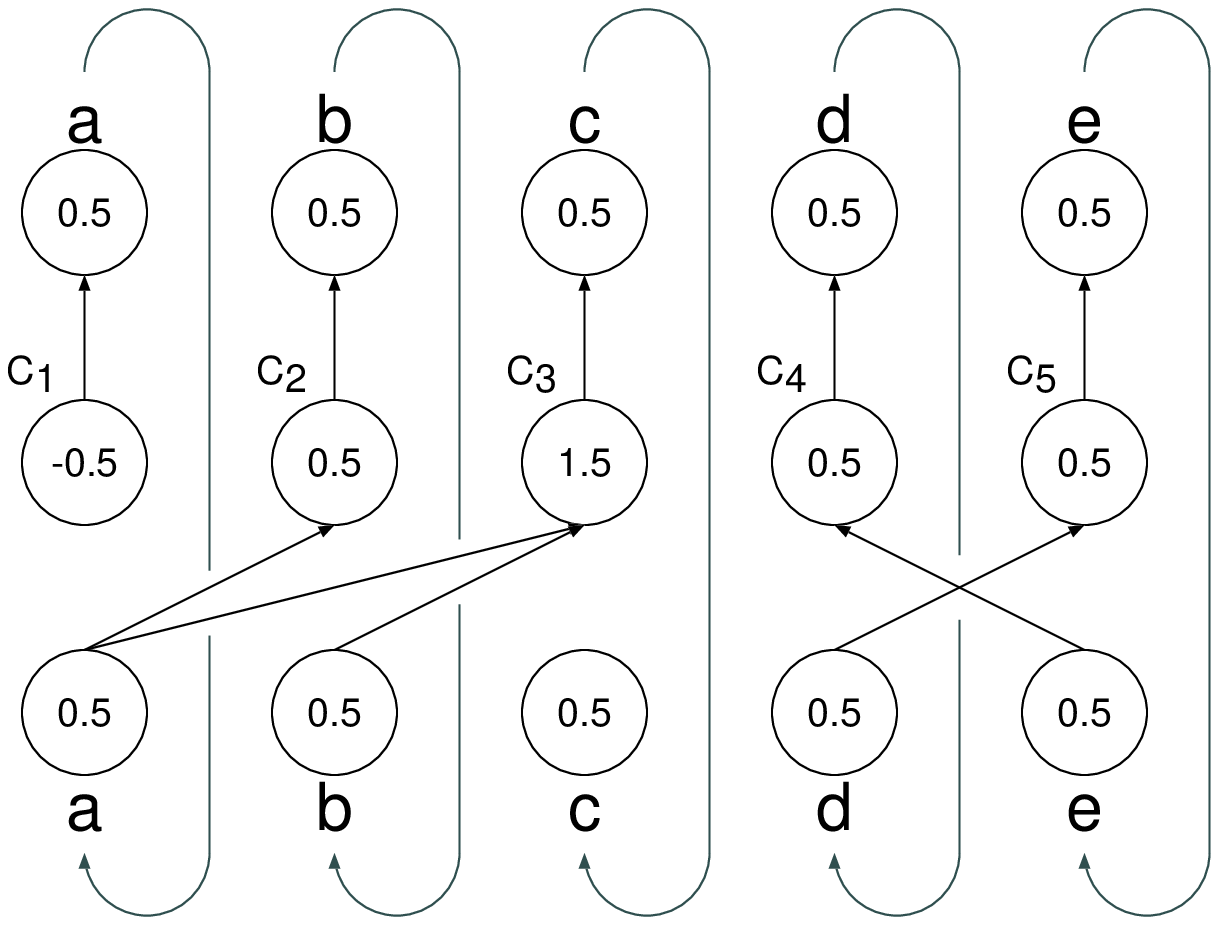, width=\linewidth}
    \end{minipage}
    \caption{A simple propositional logic program and the corresponding network. 
      Numbers within the units denote the thresholds. All weights are set to
      $1$.}
    \label{fig:hk94}
  \end{figure}
\end{example}

It can be shown that every logic program can be implemented using a 3-layer
network of binary threshold units, and that 2-layer networks do not suffice.  It
was also shown that under some syntactic restrictions on the programs, their
semantics could be recovered by recurrently connecting the output- and the input
layer of the network (as indicated in Figure \ref{fig:hk94}) and propagating
activation exhaustively through the resulting recurrent network. Key idea to
\cite{HK94} was to represent logic programs by means of their associated
\emph{semantic operators}, i.e. by connectionist encoding of an operator which
captures the meaning of the program, instead of encoding the program directly.
More precisely, the functional input-output behaviour of a semantic operator
$T_P$ associated with a given program $P$ is encoded by means of a feedforward
neural network $N_P$ which, when presented an encoding of some $I$ to its input
nodes, produces $T_P(I)$ at its output nodes. Output nodes can also be connected
recurrently back to the input nodes, resulting in a connectionist computation of
iterates of $I$ under $T_P$, as used e.g. in the computation of the semantics or
meaning of $P$ \cite{Llo88}. $I$, in this case, is a (Herbrand-)interpretation
for $P$, and $T_P$ is a mapping on the set $I_P$ of all
(Herbrand-)interpretations for $P$.
        
This idea for the representation of logic programs spawned several
investigations in different directions. As \cite{HK94} employed binary threshold
units as activation functions of the network nodes, the results were lifted to
sigmoidal and hence differentiable activation functions in \cite{AZC97,AZ99}.
This way, the connectionist representation of logic programs resulted in a
network architecture which could be trained using standard backpropagation
algorithms. The resulting connectionist inductive learning and reasoning system
CILP was completed by providing corresponding knowledge extraction algorithms
\cite{ABG01}. Further extensions to this include modal \cite{ALG02} and
intuitionistic logics \cite{Garcez:Lamb:Gabbay:2003}. Metalevel priories between
rules were introduced in \cite{ABG00}.  An in-depth treatment of the whole
approach can be found in \cite{ABG02}.  The \emph{knowledge based artificial
  neural networks (KBANN)} \cite{TS94} are closely related to this approach, by
using similar techniques to implement propositional logic formulae within neural
networks, but with a focus on learning.

Another work following up on \cite{HK94} concerns the connectionist treatment of
first-order logic programming. \cite{Seda05mfcsit} and \cite{Seda:Lane:2005}
approach this by approximating given first-order programs $P$ by finite
subprograms of the grounding of $P$.  These subprograms can be viewed as
propositional ones and encoded using the original algorithm from \cite{HK94}.
\cite{Seda05mfcsit} and \cite{Seda:Lane:2005} show that arbitrarily accurate
encodings are possible for certain programs including definite ones (i.e.
programs not containing negation as failure). They also lift their results to
logic programming under certain multi-valued logics.
        
A more direct approach to the representation of first-order logic programs based
on \cite{HK94} was pursued in
\cite{HKS99,HS00d,HHS0xjal,Hit04inap,BGH05flairs,BHW05nesy}.  The basic idea
again is to represent semantic operators $T_P: I_P\to I_P$ instead of the
program $P$ directly. In \cite{HK94} this was achieved by assigning
propositional variables to nodes, whose activations indicate whether the nodes
are true or false within the currently represented interpretation.  In the
propositional setting this is possible because for any given program only a
finite number of truth values of propositional variables plays a role -- and
hence the finite network can encode finitely many propositional variables in the
way indicated. For first-order programs, infinite interpretations have to be
taken into account, thus an encoding of ground atoms by one neuron each is
impossible as it would result in an infinite network, which is not
computationally feasible to work with.

The solution put forward in \cite{HKS99} is to employ the capability of standard
feedforward networks to propagate real numbers. The problem is thus reduced to
encoding $I_P$ as a set of real numbers in a computationally feasible way, and
to provide means to actually construct the networks starting from their
input-output behaviour. Since sigmoidal units can be used, the resulting
networks are trainable by backpropagation. \cite{HKS99} spelled
out these ideas in a limited setting for a small class of programs, and was
lifted in \cite{HS00d,HHS0xjal} to a more general setting, including the
treatment of multi-valued logics.  \cite{Hit04inap} related the results to logic
programming under non-monotonic semantics. In these reports, it was shown that
approximation of logic programs by means of standard feedforward networks is
possible up to any desired degree of accuracy, and for fairly general classes of
programs.  However, no algorithms for practical generation of approximating
networks from given programs could be presented. This was finally done in
\cite{BHW05nesy}, and implementations of the approach are currently under way,
and shall yield a first-order integrated neural-symbolic system with similar
capabilities as the propositional system CILP.

There exist two alternative approaches to the representation of first-order
logic programs via their semantic operators, which have not been studied in more
detail yet. The first approach, reported in \cite{BH0xjal}, uses insights from
fractal geometry as in \cite{Bar93} to construct iterated function systems whose
attractors correspond to fixed points of the semantic operators. The second
approach builds on Gabbay's \emph{Fibring logics} \cite{Gab_fibre}, and the
corresponding Fibring Neural Networks \cite{GG04}. The resulting system,
presented in \cite{BGH05flairs}, employs the fibring idea to control the firing
of nodes such that it corresponds to term matching within a logic programming
system. It is shown that certain limited kinds of first-order logic programs can
be encoded this way, such that their models can be computed using the network.


\section{A New Classification Scheme}\label{sec:classification}

\begin{figure}[tbh]
  \centering
  \epsfig{figure=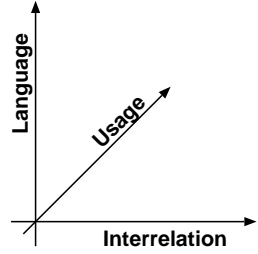, width=.3\linewidth}
  \caption{Main Axes}
  \label{fig:schema}
\end{figure}

In this section we will introduce a classification scheme for neural-symbolic
systems. This way, we intend to bring some order to the heterogeneous field of
research, whose individual approaches are often largely incomparable.  We
suggest to use a scheme consisting of three main axes as depicted in
Figure~\ref{fig:schema}, namely \emph{Interrelation}, \emph{Language} and
\emph{Usage}.

\begin{figure}[tbh]
  \epsfig{figure=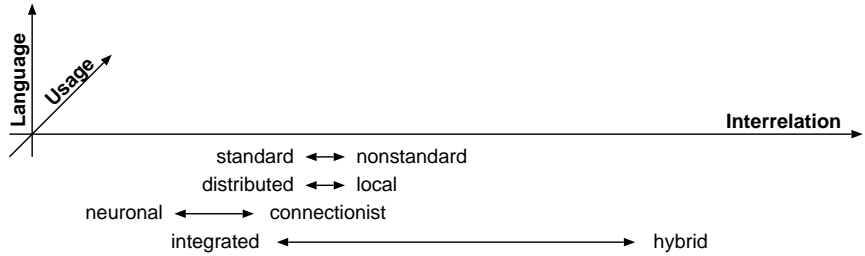, width=\linewidth}
  \caption{Interrelation}
  \label{fig:schema:integration}
\end{figure}

For the interrelation-axis, depicted in Figure~\ref{fig:schema:integration}, we
roughly follow the scheme introduced and discussed in
\cite{Hilario:1995,Hatzilygeroudis:Prentzas:2004}, but adapted to the particular
focus which we will put forward. In particular, the classifications presented in
\cite{Hilario:1995,Hatzilygeroudis:Prentzas:2004} strive to depict each system
at exactly one point in a taxonomic tree. From our perspective, certain
properties or design decisions of systems are rather independent, and should be
understood as different \emph{dimensions}. From this perspective approaches can
first be divided into two main classes, namely into \emph{integrated} (called
\emph{unified} or \emph{translational} in
\cite{Hilario:1995,Hatzilygeroudis:Prentzas:2004}) and \emph{hybrid} systems.
Integrated are those, where full symbolic processing functionalities emerge from
neural structures and processes -- further details will be discussed in
Section~\ref{sec:unifiedVShybrid}. Integrated systems can be further subdivided
into neuronal and connectionist approaches, as discussed in
Section~\ref{sec:neuronalVSconnectionist}. Neuronal indicates the usage of
neurons which are very closely related to biological neurons. In connectionist
approaches there is no claim to neurobiological plausibility, instead general
artificial neural network architectures are used.  Depending on their
architecture, they can be split into standard and non-standard networks.
Furthermore, we can distinguish local and distributed representation of the
knowledge which will also be discussed in more detail in
Section~\ref{sec:localVSdistributed}.

Note that the subdivisions belonging to the interrelation axis are again
independent of each other. They should be understood as independent
subdimensions, and could also be depicted this way by using further coordinate
axes. We hope that our simplified visualisation makes it easier to maintain an
overview. But to be pedantic, for our presentation we actually understand the
neural-connectionist dimension as a subdivision of integrated systems, and the
distributed-local and standard-nonstandard dimensions as independent
subdivisions of connectionist systems -- simply because this currently suffices
for classification.

\begin{figure}[tbh]
  \epsfig{figure=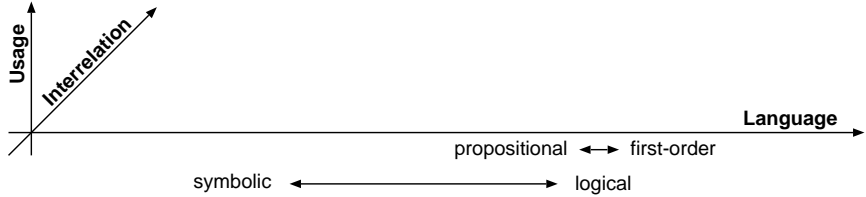, width=\linewidth}
  \caption{Language}
  \label{fig:schema:language}
\end{figure}

Figure~\ref{fig:schema:language} depicts the second axis in our scheme. Here,
the systems are divided according to the language used in their symbolic part.
We distinguish between \emph{symbolic} and \emph{logical} languages.  Symbolic
approaches include the relation to automata as in \cite{CP43}, to grammars
\cite{elman:90,Fletcher:2000:a} or to the storage and retrieval of terms
\cite{pollack:90}, whereas the logical approaches require either propositional
or first order logic systems, as e.g. in \cite{HK94} and discussed in Section
\ref{sec:HK94etc}. The language axis will be discussed in more detail in Section
\ref{sec:language}.

\begin{figure}[tbh]
  \epsfig{figure=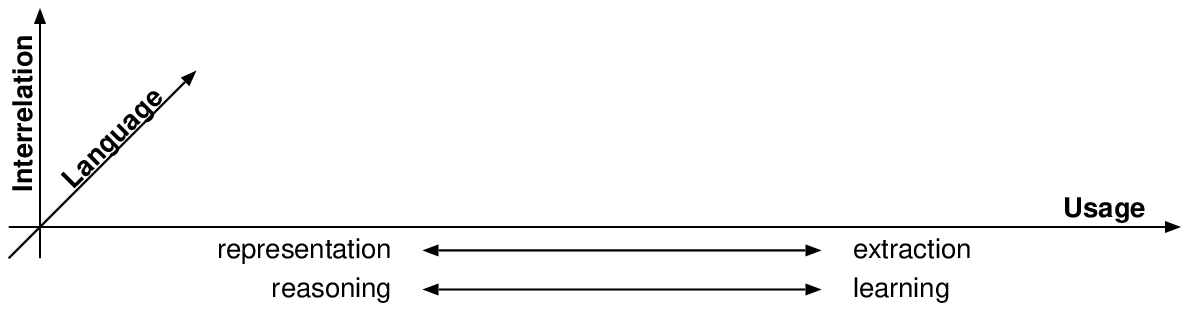, width=\linewidth}
  \caption{Usage}
  \label{fig:schema:usage}
\end{figure}

Most systems focus on one or only a few aspects of the neural-symbolic learning
cycle depicted in Figure~\ref{fig:nesycycle}, i.e. either the representation of
symbolic knowledge within a connectionist setting, or the training of
preinitialized networks, or the extraction of symbolic systems from a network.
Depending on this main focus we can distinguish the systems as shown in
Figure~\ref{fig:schema:usage}.  The issues of \emph{extraction} vs.
\emph{representation} on the one hand and \emph{learning} vs. \emph{reasoning}
on the other hand, are discussed in
Section~\ref{sec:extractionVSrepresentation}. Systems may certainly cover
several or all of these aspects, i.e. they may span whole subdimensions.

\section{Dimensions of Neural Symbolic Integration}
\label{sec:dimensions}

In this section, we will survey main research results in this area by
classifying them according to eight dimensions, marked by the arrows in
Figures~\ref{fig:schema:integration}-\ref{fig:schema:usage}.
\begin{itemize}
\item Interrelation
  \begin{enumerate}
  \item Integrated versus hybrid
  \item Neuronal versus connectionist
  \item Local versus distributed
  \item Standard versus nonstandard
  \end{enumerate}
\item Language
  \begin{enumerate}\setcounter{enumi}{4}
  \item Symbolic versus logical
  \item Propositional versus first-order
  \end{enumerate}
\item Usage
  \begin{enumerate}\setcounter{enumi}{6}
  \item Extraction versus representation
  \item Learning versus reasoning
  \end{enumerate}
\end{itemize}
As discussed above, we believe that these dimensions mark the main points of
distinction between different integrated neural-symbolic systems. The chapter is
structured accordingly, examining each of the dimensions in turn.

\subsection{Interrelation}

\subsubsection{Integrated versus Hybrid}
\label{sec:unifiedVShybrid}

This section serves to further clarify what we understand by
\emph{neural-symbolic integration}. Following the rationale laid out in the
introduction, we understand why it is desirable to combine symbolic and
connectionist approaches, and there are obviously several ways how this can be
done. From a bird's eye view, we can distinguish two main paradigms, which we
call \emph{hybrid} and \emph{integrated} (or following \cite{Hilario:1995},
\emph{unified}) systems, and this survey is concerned with the latter.

\begin{figure}
  \centering
  \begin{minipage}{.5\linewidth}
    \epsfig{figure=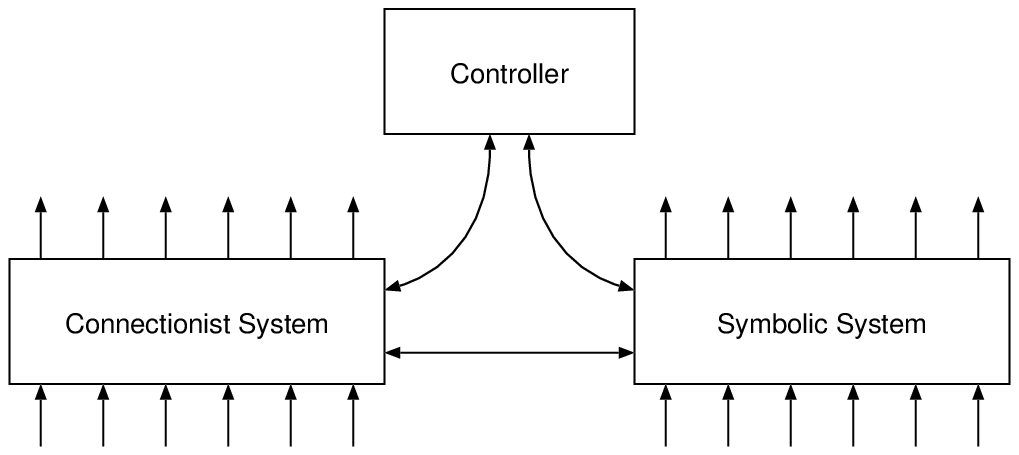, width=\linewidth}
  \end{minipage}
  \hspace{.1\linewidth}
  \begin{minipage}{.3\linewidth}
    \footnotesize
    \centering
    $Can\!\!-\!\!Sell(josephine, ball)$\\
    \epsfig{figure=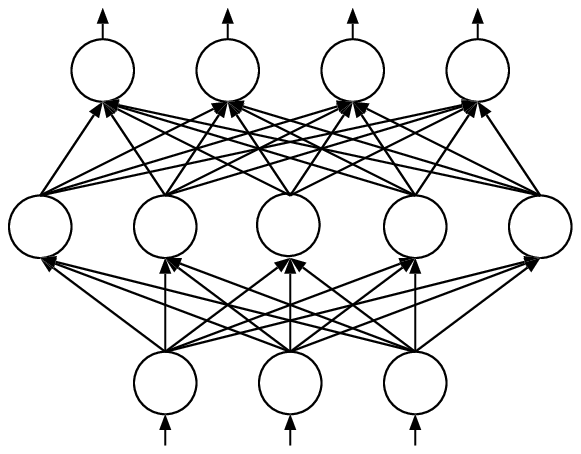, width=\linewidth}
    $Onws(josephine, ball)$
  \end{minipage}
  \caption{Hybrid (left) versus integrated (right) architecture.}
  \label{fig:hybridVSIntegrated}
\end{figure}

\emph{Hybrid} systems are characterized by the fact that they combine two or
more problem-solving techniques in order to address a problem, which run in
parallel, as depicted in Figure~\ref{fig:hybridVSIntegrated}.

An \emph{integrated} neural-symbolic system differs from a hybrid one in that it
consists of one connectionist main component in which symbolic knowledge is
processed, see Figure \ref{fig:hybridVSIntegrated} (right).  Integrated systems
are sometimes also referred to as \emph{embedded} or \emph{monolithic} hybrid
systems, cf. \cite{SunECS01}. Examples for integrated systems are e.g. those
presented in Sections \ref{sec:raams}-\ref{sec:HK94etc}.

For either architecture, one of the central issues is the representation of
symbolic data in connectionist form \cite{BHH04}. For the hybrid system, these
transformations are required for passing information between the components. The
integrated architecture must implicitly or explicitly deal with symbolic data by
connectionist means, i.e. must also be capable of similar transformations.

This survey covers integrated systems only, the study of which appears to be
particularly challenging.  For recent selective overview literature see e.g.
\cite{BS01,ABG02,BHH04}. The first, \cite{BS01}, focuses on reasoning systems.
The field of propositional logic is thoroughly covered in \cite{ABG02}, where
the authors revisit the approach of \cite{HK94} and explain their extensions
including applications to real world problems, like fault diagnosis. In
\cite{BHH04} the emphasis is on the challenge problems arising from first-order
neural-symbolic integration.

\subsubsection{Neuronal versus Connectionist}
\label{sec:neuronalVSconnectionist}

There are two driving forces behind the field of neural-symbolic integration: On
the one hand it is the striving for an understanding of human cognition, and on
the other it is the vision of combining connectionist and symbolic artificial
intelligence technology in order to arrive at more powerful reasoning and
learning systems for computer science applications.
        
In \cite{CP43} the motivation for the study was to understand human cognition,
i.e. to pursue the question how higher cognitive -- logical -- processes can be
performed by artificial neural networks. In this line of research, the question
of \emph{biological feasability} of a network architecture is prominent, and
inspiration is often taken from biological counterparts.
        
The SHRUTI system \cite{SA93} as described in Section~\ref{sec:shruti}, for
example, addresses the question how it is possible that biological networks
perform certain reasoning tasks very quickly.  Indeed, for some complex
recognition tasks which involve reasoning capabilities, human responses occur
sometimes at \emph{reflexive} speed, particularly within a time span which
allows processing through very few neuron layers only.  As mentioned above,
time-synchronization was used for the encoding of variable binding in SHRUTI.

The recently developed spiking neurons networks \cite{MaassParadigmsSpiking02}
take an even more realistic approach to the modelling of temporal aspects of
neural activity. Neurons, in this context, are considered to be firing so-called
\emph{spike trains}, which consist of patterns of firing impulses over certain
time intervals. The complex propagation patterns within a network are usually
analysed by statistical methods. The encoding of symbolic knowledge using such
temporal aspects has hardly been studied so far, an exception being
\cite{Sougne:2001}. We perceive it as an important research challenge to relate
the neurally plausible spiking neurons approach to neural-symbolic integration
research. To date, however, only a few preliminary results on computational
aspects of spiking neurons have been obtained
\cite{NatschlaegerMaass:2001,MaassMarkram:02,MaassETAL:02c}.

Another recent publication, \cite{VeldeKampsBBS05}, shows how natural language
could be encoded using biologically plausible models of neural networks. The
results appear to be suitable for the study of neural-symbolic integration, but
it remains to be investigated to which extent the provided approach can be
transfered to symbolic reasoning. Similarly inspiring might be the recent book
\cite{JeffHawkins04book} and accompanying work, though it discusses
neural-symbolic relationships on a very abstract level only.
        
The lines of research just reviewed take their major motivation from the goal to
achieve biologically plausible behaviour or architectures. As already mentioned,
neural-symbolic integration can also be pursued from a more technically
motivated perspective, driven by the goal to combine the advantages of symbolic
and connectionist approaches by studying their interrelationships.  The work on
the Core Method, discussed in Section \ref{sec:HK94etc}, can be subsumed under
this technologically inspired perspective.

\subsubsection{Local versus Distributed Representation of Knowledge}
\label{sec:localVSdistributed}

For integrated neural-symbolic systems, the question is crucial how symbolic
knowledge is represented within the connectionist system. If standard networks
are being trained using backpropagation, the knowledge acquired during the
learning process is spread over the network in diffuse ways, i.e. it is in
general not easy or even possible to identify one or a small number of nodes
whose activations contain and process a certain symbolic piece of knowledge.

The RAAM architecture and their variants as discussed in Section \ref{sec:raams}
are clearly based on distributed representations. Technically, this stems from
the fact that the representation is initially learned, and no explicit algorithm
for translating symbolic knowledge into the connectionist setting is being used.

Most other approaches to neural-symbolic integration, however, represent data
locally. SHRUTI (Section \ref{sec:shruti}) associates a defined node assembly to
each logical predicate, and the architecure does not allow for distributed
representation. The approaches for propositional connectionist model generation
using the Core Method (Section \ref{sec:HK94etc}) encode propositional variables
as single nodes in the input resp. output layer, and logical formulae (rules) by
single nodes in the hidden layer of the network.

The design of distributed encodings of symbolic data appears to be particular
challenging. It also appears to be one of the major bottlenecks in producing
applicable integrated neural-symbolic systems with learning and reasoning
abilities \cite{BHH04}. This becomes apparent e.g. in the difficulties faced by
the first-order logic programming approaches discussed in Section
\ref{sec:HK94etc}. Therein, symbolic entities are not represented directly.
Instead, interpretations (i.e. valuations) of the logic are being represented,
which contain truth value assignments to language constructs. Concrete
representations, as developed in \cite{BHW05nesy}, distribute the encoding of
the interpretations over several nodes, but in a diffuse way. The encoding thus
results in a distributed representation. Similar considerations apply to the
recent proposal \cite{Gust:Kuehnberger:2005}, where first-order logic is first
converted into variable-free form (using topoi from category theory), and then
fed to a neural network for training.

\subsubsection{Standard versus Non/standard Network Architecture}
\label{sec:standardVSnonstandard}

Even though neural networks are a widely accepted paradigm in AI it is hard to
make out a standard architecture. But, all so called standard-architecture
systems agree at least on the following:
\begin{itemize}
\item only real numbers are propagated along the connections
\item units compute very simple functions only
\item all units behave similarly (i.e. they use similar simple functions and the
  activation values are always within a small range)
\item only simple recursive structures are used (e.g. connecting only the output
  back to the input layer, or use selfrecursive units only)
\end{itemize}
When adhering to these standard design principles, powerful learning techniques
as e.g. backpropagation \cite{RHW86} or Hebbian Learning \cite{Hebb:1949} can be
used to train the networks, which makes them applicable to real world problems.

However, these standard architectures do not easily lend themselves to
neural-symbolic integration. In general, it is easier to use non-standard
architectures in order to represent and work with structured knowledge, with the
drawback that powerful learning abilities are often lost.

Neural-symbolic approaches using standard networks are e.g. the CILP system
\cite{AZ99}, KBANN \cite{TS94}, RAAM (Section \ref{sec:raams}) and
\cite{Seda:Lane:2005} (Section \ref{sec:HK94etc}).  Usually, they consist of a
layered network, consisting of three or in case of KBANN more layers, and
sigmoidal units are being used. For these systems experimental results are
available showing their learning capabilities. As discussed above, these systems
are able to handle propositional knowledge (or first order with a finite
domain). Similar observations can be made about the standard architectures used
in \cite{HKS99,HHS0xjal,BHW05nesy} for first-order neural-symbolic integration.

Non-standard networks were used e.g. in the SHRUTI system \cite{SA93}, and in
the approaches described in \cite{BH0xjal} and \cite{BGH05flairs}. In all these
implementations non-standard units and non-standard architectures were used, and
hence none of the usual learning techniques are applicable. However, for the
SHRUTI system limited learning techniques based on Hebbian Learning
\cite{Hebb:1949} were developed
\cite{Shastri:2002,Shastri:Wendelken:2003,WS03pcs}.

\subsection{Language}\label{sec:language}

\subsubsection{Symbolic versus Logical}

One of the motivations for studying neural-symbolic integration is to combine
connectionist learning capabilities with symbolic knowledge processing, as
already mentioned. While our main interest is in pursuing logical aspects of
symbolic knowledge, this is not necessarily always the main focus of
investigations.

Work on representing automata or weighted automata \cite{KleeneAS56,CP43,BHS04}
(Section \ref{sec:cp43}) using artificial neural networks, for example focuses
on computationally relevant structures, such as automata, and not directly on
logically encoded knowledge. Nevertheless, such investigations show how to deal
with structural knowledge within a connectionist setting, and can serve as
inspiration for corresponding research on logical knowledge.

Recursive autoassociative memory, RAAM, and their variants as discussed in
Section \ref{sec:raams}, deals with terms only, and not directly with logical
content. RAAM allows connectionist encodings of first-order terms, where the
underlying idea is to present terms or term trees sequentially to a
connectionist system which is trained to produce a compressed encoding
characterized by the activation pattern of a small collection of nodes. To date,
storage capacity is very limited, and connectionist processing of the stored
knowledge has not yet been investigated in detail.

A considerable body of work exists on the connectionist processing and learning
of structured data using recurrent networks
\cite{SSG95,SSG97,FGKS01sequences,HammerCSR02,HammerAaL03,HMSS04nn,HMSS04nc}.
The focus is on tree representations and manipulation of the data.

\cite{hkl:97,kalinke:lehmann:98} study the representation of counters using
recurrent networks, and connectionist unification algorithms as studied in
\cite{h:90:b,hk:92,Hol93} are designed for manipulating terms, but already in a
clearly logical context. The representation of grammars \cite{Giles:ETAL:1991}
or more generally of natural language constructs \cite{VeldeKampsBBS05} also has
a clearly symbolic (as opposed to logical) focus.

It remains to be seen, however, to what extent the work on connectionist
processing of structured data can be reused in logical contexts for creating
integrated neural-symbolic systems with reasoning capabilities. Integrated
reasoning systems like the ones presented in Sections \ref{sec:shruti} and
\ref{sec:HK94etc} currently lack the capabilities of the term-based systems, so
that a merging of these efforts appears to be a promising albeit challenging
goal.

\subsubsection{Propositional versus First-Order}

Logic-based integrated neural-symbolic systems differ as to the knowledge
representation language they are able to represent. Concerning the capabilities
of the systems, a major distinction needs to be made between those which deal
with propositional logics, and those based on first-order predicate (and
related) logics.

What we mean by propositional logics in this context includes propositional
modal, temporal, non-monotonic, and other non-classical logics. One of their
characteristic feature which distinguishes them from first-order logics for
neural-symbolic integration is the fact that they are of a finitary nature:
propositional theories in practice involve only a finite number of propositional
variables, and corresponding models are also finite. Also, sophisticated symbol
processing as needed for nested terms in the form of substitutions or
unification is not required.

Due to their finiteness it is thus fairly easy to implement propositional logic
programs using neural networks \cite{HK94} (Section \ref{sec:HK94etc}). A
considerable body of work deals with the extension of this approach to
non-classical logics
\cite{Garcez:Gabbay:Lamb:2005,ABG00,ALG02,Garcez:Lamb:Gabbay:2003,Garcez:Gabbay:Lamb:2004,GLBG04jait,GGL0xjlc}.
This includes modal, intuitionistic, and argumentation-theoretic approaches,
amongst others.  Earlier work on representing propositional logics is based on
Hopfield networks \cite{pinkas:91:a,pinkas:91} but has not been followed up on
recently.

In contrast to this, predicate logics -- which for us also include modal,
non-monotonic, etc. extensions -- in general allow to use function symbols as
language primitives. Consequently, it is possible to use terms of arbitrary
depth, and models necessarily assign truth values to an infinite number of
ground atoms. The difficulty in dealing with this in a connectionist setting
lies in the finiteness of neural networks, necessitating to capture the
infinitary aspects of predicate logics by finite means. The first-order
approaches presented in
\cite{HKS99,HS00d,BH0xjal,HHS0xjal,BGH05flairs,BHW05nesy} (Section
\ref{sec:HK94etc}) solve this problem by using encodings of infinite sets by
real numbers, and representing them in an approximate manner. They can also be
carried over to non-monotonic logics \cite{Hit04inap}.

\cite{BGH05flairs}, which builds on \cite{GG04} and \cite{Gab_fibre} uses an
alternative mechanism in which unification of terms is controlled via fibrings.
More precisely, certain network constructs encode the matching of terms and act
as gates to the firing of neurons whenever corresponding symbolic matching is
achieved.

A prominent subproblem in first-order neural-symbolic integration is that of
variable binding. It refers to the fact that the same variable may occur in
several places in a formula, or that during a reasoning process variables may be
bound to instantiate certain terms. In a connectionist setting, different parts
of formulae and different individuals or terms are usually represented
independently of each other within the system. The neural network paradigm,
however, forces subnets to be blind with respect to detailed activation patterns
in other subnets, and thus does not lend itself easily to the processing of
variable bindings.

Research on first-order neural-symbolic integration has led to different means
of dealing with the variable binding problem. One of them is to use temporal
synchrony to achieve the binding. This is encoded in the SHRUTI system (Section
\ref{sec:shruti}), where the synchronous firing of variable nodes with constant
nodes encodes a corresponding binding. Other approaches, as discussed in
\cite{Browne:Sun:1999}, encode binding by relating the propagated activations,
i.e. real numbers.

Other systems avoid the variable binding problem by converting predicate logical
formulae into variable-free representations. The approaches in
\cite{HKS99,HS00d,HHS0xjal,Hit04inap,Seda05mfcsit,Seda:Lane:2005,BGH05flairs,BHW05nesy}
(Section \ref{sec:HK94etc}) make conversions to (infinite) propositional
theories, which are then approximated. \cite{Gust:Kuehnberger:2005} use topos
theory instead.

It shall be noted here that SHRUTI (Section \ref{sec:shruti}) addresses the
variable binding problem, but allows to encode only a very limited fragment of
first-order predicate logic \cite{HKW99}. In particular, it does not allow to
deal with function symbols, and thus could still be understood as a finitary
fragment of predicate logic.

\subsection{Usage}

\subsubsection{Extraction versus Representation}
\label{sec:extractionVSrepresentation}

The representation of symbolic knowledge is necessary even for classical
applications of connectionist learning. As an example, consider the
neural-networks-based Backgammon playing program TD-Gammon \cite{Tesauro95cacm},
which achieves professional players' strength by temporal difference learning on
data created by playing against itself. TD-Gammon represents the Backgammon
board in a straightforward way, by encoding the squares and placement of pieces
via assemblies of nodes, thus representing the structured knowledge of a board
situation directly by a certain activation pattern of the input nodes.

In this and other classical application cases the represented symbolic knowledge
is not of a complex logical nature. Neural-symbolic integration, however,
attempts to achieve connectionist processing of complex logical knowledge,
learning, and inferences, and thus the question how to represent logical
knowledge bases in suitable form becomes dominant. Different forms of
representation have already been discussed in the context of local versus
distributed representations.

Returning to the TD-Gammon example, we would also be interested in the complex
knowledge as acquired by TD-gammon during the learning process, encoding the
strategies with which this program beats human players. If such knowledge could
be extracted in symbolic form, it could be used for further symbolic processing
using inference engines or other knowledge based systems.

It is apparent, that both the representation and the extraction of knowledge
are of importance for integrated neural-symbolic systems. They are needed for
closing the neural-symbolic learning cycle (Figure \ref{fig:nesycycle}).
However, they are also of independent interest, and are often studied
separately. 

As for the representation of knowledge, this component is present in all systems
presented so far. The choice how representation is done often determines whether
standard architectures are used, if a local or distributed approach is taken,
and whether standard learning algorithms can be employed.

A large body of work exists on extracting knowledge from trained networks,
usually focusing on the extraction of rules. \cite{Jacobson:2005} gives a
recent overview over extraction methods.  A method from 1992
\cite{Giles:ETAL:1991} is still up to date, where a method is given to extract a
grammar represented as a finite state machine from a trained recurrent neural
network. \cite{McGarray:ETAL:1999} show how to extract rules from radial basis
function networks by identifying minimal and maximal activation values. Some of
the other efforts are reported in
\cite{TS93,ADT95,Bologna:2000,ABG01,LehmannBaderHitzler:2005}

It shall be noted that only a few systems have been proposed to date which
include representation, learning, and extraction capabilities in a meaningful
way, one of them being CILP \cite{AZC97,AZ99,ABG01}. It is to date a difficult
research challenge to provide similar functionalities in a first-order setting.

\subsubsection{Learning versus Reasoning}
\label{sec:learningVSreasoning}

Ultimately, our goal should be to produce an effective AI system with added
reasoning and learning capabilities, as recently pointed out by Valiant
\cite{Valiant:2003} as a key challenge for computer science. It turns out that
most current systems have either learning capabilities or reasoning
capabilities, but rarely both. SHRUTI (Section \ref{sec:shruti}), for example,
is a reasoning system with very limited learning support. 

In order to advance the state of the art in the sense of Valiant's vision
mentioned above, it will be necessary to install systems with combined
capabilities. In particular, learning should not be independent of reasoning,
i.e. initial knowledge and logical consequences thereof should help guiding the
learning process. There is no system to-date which realizes this in any way, and
new ideas will be needed to attack this problem.


\section{Conclusions and Further Work}\label{sec:conclusions}

Intelligent systems based on symbolic knowledge processing, on the one hand, and
on artificial neural networks, on the other, differ substantially. Nevertheless,
these are both standard approaches to artificial intelligence and it would be
very desirable to combine the robustness of neural networks with the
expressivity of symbolic knowledge representation.  This is the reason why the
importance of the efforts to bridge the gap between the connectionist and
symbolic paradigms of Artificial Intelligence has been widely recognised. As the
amount of hybrid data containing symbolic and statistical elements as well as
noise increases in diverse areas such as bioinformatics or text and web mining,
neural-symbolic learning and reasoning becomes of particular practical
importance. Notwithstanding, this is not an easy task, as illustrated in the
survey.

The merging of theory (background knowledge) and data learning (learning from
examples) in neural networks has been indicated to provide learning systems that
are more effective than e.g. purely symbolic and purely connectionist systems,
especially when data are noisy \cite{AZ99}. This has contributed decisively to
the growing interest in developing neural-symbolic systems, i.e. hybrid systems
based on neural networks that are capable of learning from examples and
background knowledge, and of performing reasoning tasks in a massively parallel
fashion.

However, while symbolic knowledge representation is highly recursive and well
understood from a declarative point of view, neural networks encode knowledge
implicitly in their weights as a result of learning and generalisation from raw
data, which are usually characterized by simple feature vectors. While
significant theoretical progress has recently been made on knowledge
representation and reasoning using neural networks, and on direct processing of
symbolic and structured data using neural methods, the integration of neural
computation and expressive logics such as first order logic is still in its
early stages of methodological development.

Concerning knowledge extraction, we know that neural networks have been applied
to a variety of real-world problems (e.g. in bioinformatics, engineering,
robotics), and they were particularly successful when data are noisy. But
entirely satisfactory methods for extracting symbolic knowledge from such
trained networks in terms of accuracy, efficiency, rule comprehensibility, and
soundness are still to be found. And problems on the stability and learnability
of recursive models currently impose further restrictions on connectionist
systems.

In order to advance the state of the art, we believe that it is necessary to
look at the biological inspiration for neural-symbolic integration, to use more
formal approaches for translating between the connectionist and symbolic
paradigms, and to pay more attention to potential application scenarios.

The general motivation for research in the field of neural-symbolic integration
(just given) arises from conceptual observations on the complementary nature of
symbolic and neural network based artificial intelligence described above. This
conceptual perspective is sufficient for justifying the mainly
foundations-driven lines of research being undertaken in this area so far.
However, it appears that this conceptual approach to the study of
neural-symbolic integration has now reached an impasse which requires the
identification of use cases and application scenarios in order to drive future
research.

Indeed, the theory of integrated neural-symbolic systems has reached a quite
mature state but has not been tested extensively so far on real application
data.  The current systems have been developed for the study of general
principles, and are in general not suitable for real data or application
scenarios that go beyond propositional logic. Nevertheless, these studies
provide methods which can be exploited for the development of tools for use
cases, and significant progress can now only be expected as a continuation of
the fundamental research undertaken in the past.

In particular, first-order neural-symbolic integration still remains a widely
open issue, where advances are very difficult, and it is very hard to judge to
date to what extent the theoretical approaches can work in practice. We believe
that the development of use cases with varying levels of expressive complexity
is, as a result, needed to drive the development of methods for neural-symbolic
integration beyond propositional logic \cite{HBG05nesy}.

\nocite{Barnden:1995,Healy:1999,Sougne:2001,Niklasson:Linaker:2000,Rodriguez:1999,Gallant:1993,Hatzilygeroudis:Prentzas:2000,McGarray:ETAL:1999}

\end{document}